# Unification of multi-lingual scientific terminological resources using the ISO 16642 standard. The TermSciences initiative

**Majid Khayari[1], Stéphane Schneider[1], Isabelle Kramer[2], Laurent Romary[2]**

INIST-CNRS

2, Allée du Parc de Brabois- CS 10310

54519 Vandoeuvre-Lès-Nancy

LORIA, Universities NANCY, CNRS INRIA

Campus Scientifique -BP 239

54506 Vandoeuvre-Lès-Nancy-Cedex

[majid.khayari@inist.fr](mailto:majid.khayari@inist.fr), [stephane.schneider@inist.fr](mailto:stephane.schneider@inist.fr), [isabelle.kramer@loria.fr](mailto:isabelle.kramer@loria.fr), [laurent.romary@loria.fr](mailto:laurent.romary@loria.fr)

**Abstract**

The TermSciences initiative aims at building a multi-purpose and multi-lingual knowledge system from different source vocabularies produced by major French research institutions and which were initially intended to be used for indexing and cataloguing scientific literature. Since the construction of language resource repositories is cost-effective and time-consuming, the producers of these vocabularies wished to both share their terminological material and develop common tools for the collaborative management of the integrated resource. Sharing terminologies poses some problems because of the heterogeneous nature of the source data (i.e., coverage, granularity and compositionality of concepts, etc.), and to the discrepancy between partner needs (i.e., simple diffusion of the terminological material, use of the shared material to enhance information engineering tasks, etc.). This paper presents the TermSciences portal[1], which deals with the implementation of a conceptual model that uses the recent ISO 16642 standard (Terminological Markup Framework). This standard turned out to be suitable for concept modeling since it allowed for organizing the original resources by concepts and to associate the various terms for a given concept. Additional structuring is produced by sharing conceptual relationships, that is, cross-linking of resource results through the introduction of semantic relations which may have initially be missing. A special emphasis is put on medical resources used in this project, i.e. the French translation by the Institut National de la Santé et de la Recherche Médicale (INSERM) of the MeSH thesaurus from the US National Library of Medicine, the public health thesaurus of the Banque de Données de Santé Publique (BDSP) and the dictionary of human and mammals reproduction biotechnology of the Institut National de la Recherche Agronomique (INRA).

## 1. INTRODUCTION

The development of Communication and Information Technologies, and in particular, in the field of natural language resources, including terminology, raises the crucial question of standardization. Since the construction of language resource repositories is cost-effective and time-consuming, the producers and users of specialized vocabularies may benefit from sharing their resources. Still, sharing resources implies to agree about common formats and data models. This paper presents the TermSciences initiative whose purpose is to build a common terminological reference database (Bourigault and Condamines, 1995) from terminological resources (lexicons, dictionaries, thesauri) produced and maintained by various French public research institutions. As such, it is the first public initiative to implement the recently adopted Terminological Markup Framework (TMF, ISO 16642). TMF aims at providing a platform for the interchange of computerized lexical data, as used in many kinds of applications.

In this context, an important issue is to provide a uniform way of representing such databases considering the heterogeneity of both their formats and their descriptors. This is an essential aspect of natural language processing since it allows for both reusing linguistic data such as lexicons or grammars and deploying interoperable linguistic components in complex processing lines. The TermSciences project allowed us to validate step by step different stages related to the deployment of such an infrastructure, within the context of a concrete implementation of the TMF methodology and principles: modelling (ISO 704), import, fusion, update and export of data, and modification of the model.

## 2. REQUIREMENTS

### 2.1. Need of conceptualization

A major obstacle to the sharing of terminologies is the lack of conceptual integration of terms (Gangemi and al, 1998). Since the meaning of terms may be different according to the domain in which they appear (Wüster, 1976) and to the context of use (Rastier, 1995), any successful integration relies on a conceptualization process. However, most terminologies used in this project were built according to a term-centred (i.e. a descriptor-oriented) model (Condamines, 1994). This means that the linking of terms to concepts implies firstly to find or define some abstract high level terminologies (list of concepts) or ontologies and then to clear and consensual definition of concepts, i.e. if multiple terms (synonyms) may refer to the same object, a concept is unique for a given object and there is no place for an alternate or complimentary concept related to the same object (Baud et al, 1998).

---

[1] www.termsciences.fr

### 2.2. Documenting meta data

A major issue of the TermSciences initiative is the management of the integrated terminological database. Because the common database is being built from resources managed independently by different institutions, the conceptual model includes meta data about the sources of each element composing a terminological entry. Additionally, every native resource file is formatted in the target format and stored as is. The use of pointing mechanisms based on "xml:id" and the XPointer syntax make it possible to reach any native record in these formatted files and capture new elements such us updates made lately by the producer of a given resource.

### 2.3. Collaborative Update

The management of the terminological content is planned to be taken in charge by collaborators that are involved in terminological works and by others who are indexers dealing more with indexing vocabularies (i.e. artificial languages) than with terminologies. This implies that staff education is a pre-requisite to the advancement of this project. The essential difference between words and concepts, the notion of synonymy, which applies, to the first but not to the second, and the need of a natural "compositionality" of terms represent the main distinctions to be made.

## 3. TMF

The representation using TMF can be summarized as the description of computerized terminological data representation languages; it is based on two components: a meta-model, i.e. the underlying structural skeleton and a description of constraints of attachment of some information to the structural model, i.e. data categories as described in the ISO 12620 standard.

### 3.1. TMF metamodel

A meta-model does not describe one specific format, but acts as a kind of high level mechanism based on the following elementary notions: structure, information, and methodology. The structuring elements of the meta-model are called "components" and they may be "decorated" with information units, called Data Categories. A meta-model should also comprise a flexible specification platform for elementary units. This specification platform should be coupled to a reference set of descriptors that should be used to parameterize specific applications dealing with content. The terminological meta-model is based on guidelines concerning the methods and principles of terminology management involving the production of terminological entries as described in ISO 704 (ISO 704). Because a terminology always deals with special language in a particular field of knowledge, the concept shall be viewed as a unit of knowledge. The concept is a higher level of abstraction in a terminology; it links an object and its designations. The concepts contextualized in the special language of the subject field can be expressed in the various forms: terms, appellations, definitions or other linguistic forms (ISO 704). One of the most important characteristics of a terminological entry is its concept orientation: a terminological entry represents one concept which is designated by one or several terms in one or several languages.

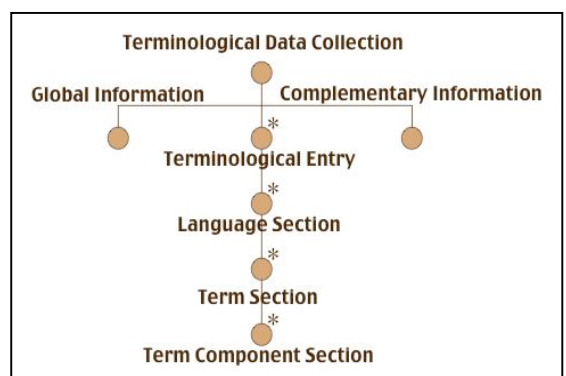

Figure 1: TMF Meta-model

Each entry can have multiple language sections, and each language section can have multiple terminological units. Each data element in an entry can be associated with various kinds of descriptive and administrative information.

### 3.2. Data category

A meta-model contains several information units related to a given format, which we refer to as "Data Categories". A selection of data categories (DCS) can be derived as a subset of a Data Category Registry (DCR) (Ide and Romary, 2004) ensuring that the semantics of these data categories are well defined and accepted by community of specialists. A data category is the generic term that references a concept. For example, the data category /*originatingInstitution*/ indicates an institution (i.e. company, government agency, etc.) treated as a source of information for the purpose of bibliographic documentation. For each element in TermSciences, the originating institution is mentioned in order to document the source of the data. A Data category Selection is needed in order to define, in combination with a meta-model, the various constraints that apply to a given domain-specific information structure or interchange format. A DCS is firstly used to specify constraints on the implementation of a meta-model instantiation, and secondly to provide the necessary information for implementing filters that convert one instantiation to another and to produce a "Generic Mapping Tool" (GMT) representation.

### 3.3. Introduction to GMT

GMT can be considered as a XML canonical representation of the generic model. The hierarchical organization of the meta-model and the qualification of each structural level can be realized in XML by instantiating the abstract structure shown above (Figure 2) and associating information units to this structure. The meta-model can be represented by means of a generic element <struct> (for structure) which can recursively express the embedding of the various representation levels of a TMF instance. Each structural node in the meta-model shall be identified by means of a type attribute associated with the <struct> element. The possible values of the type attribute shall be the identifiers of the levels in the meta-model (i.e., Terminological Data Collection, Global Information, Terminological Entry, Language Section, Term Section, Term Component Section).

Basic information units associated with a structural skeleton can be represented using the <feat> (for feature) element. Compound information units can be represented using the <brack> (for bracket) element, which can itself contain a <feat> element followed by any combination of <feat> elements and <brack> elements. Each information unit must be qualified with a type attribute, which shall take as its value the name of a standardized data category or one user-defined data category.

## 4. IMPLEMENTATION

As the source vocabularies are diverse with respect to format, structure and content, they were analyzed and restructure to fit the meta-model, in order to allow for high interoperability between terminological systems. Following this, comparisons were made between all the resources and common concepts were grouped in terminological entries in which data belonging to different resources were issued with their sources. Terminological resources

### 4.1. Terminological resources

The terminologies used in the preliminary phase of this project are vocabularies from four French research institutes: indexing vocabularies from the Institut de l'Information Scientfique et Technique (INIST-CNRS); the MeSH thesaurus from the US National Library of Medicine including its French translation by the Institut de la Santé et de la Recherche Médicale (INSERM); the thesaurus of public health produced by the Banque de données de Santé Publique (BDSP) and the Dictionary of Human and mammals reproduction biotechnology produced by the Institut National de la Recherche Agronomique (INRA).

### 4.2. From descriptors to concept

Instead of simply being aggregated, these native resources were fused together. For example, the term "*Gamete Intrafallopian Transfer*" with several translations and a definition (Figure 2) was found in NLM, INRA and INIST resources and it refers to the same object.

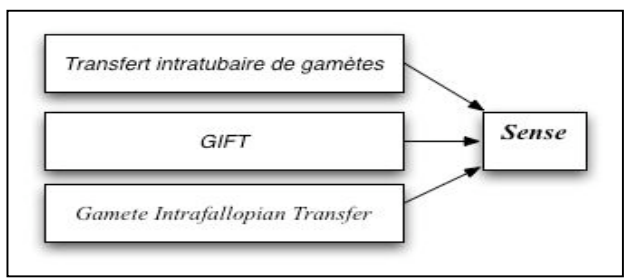

Figure 2: Semasiologic view of GIFT

Thesauri or lexicons present a semasiologic view of the world (figure 2) and are frequently arranged by alphabetic order. The main challenge of this project was to have another view of the data, no more a semasiologic view but rather an onomasiologic one (Romary andVan Campenhoudt, 2001) (Figure 3).

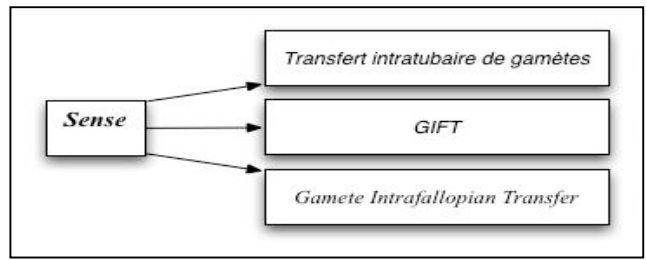

Figure 3: Onomasiologic view

### 4.3. Heterogeneous data

The resulting terminological record for a given concept presents terms and relationships that may be conflicting. In source terminologies such as the MeSH thesaurus or the public health thesaurus which are organized and used for library indexing, different concepts may be present in the same record under the same descriptor depending on the degree of specificity. For example, the record in BDSP thesaurus presents the term "Brain" as a descriptor for "Cortex", i.e. a "Used for" relation links the two terms in this thesaurus which presents only broader levels for anatomical terms. When this record was processed for integration in the common terminological database, the term "Cortex" was captured as a synonym of the term "Brain". In highly structured resources such as the MeSH thesaurus, entry terms which are synonyms, or closely related terms are documented as non-preferred concepts which allowed us to discard them during the integration process. Additionally, every resource comes up with its own categories and relationships. Thus, this first substrate needs major improvements in terms of smoothing of conflicts that may appear between concept categorization or semantic networking strategies.

In the integrated TermSciences terminological content it is important to document and identify the source of each element. Thus, the resulting terminological record for a given concept presents meta data for terms, relationships, definition, etc. These meta data allows for inclusion of some administrative information like the last modification date for an element. Additionally partners can update or export their own data according to their origin. Figure 4 illustrates the documentation of sources meta data for the above example on figure 2. The concept will be illustrated by a definition and a set of terms in different languages; each element being accompanied by its origin.

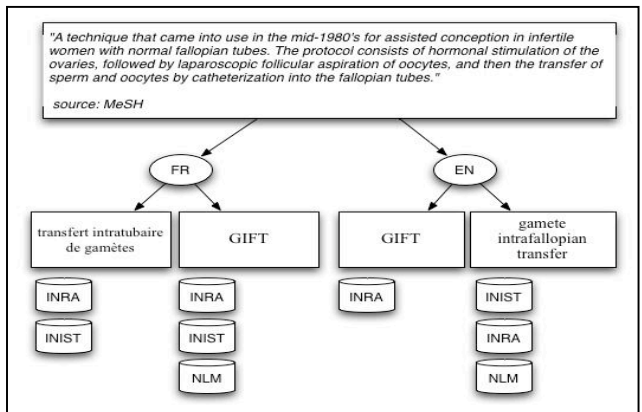

Figure 4: Data sources

The meta data shows the origin institution and/or database, but can also give a bibliographical reference. For example, several partners furnished this term "Gamete Intrafallopian Transfer"; one of them published this vocabulary. It is important to be able to complete institutional information by a bibliographical source (figure 5).

```
<struct type="languageSection">
  <feat type="languageIdentifier">en</feat>
  <brack>
    <feat type="context">
      <annot type="term">Gamete intrafallopian transfer</annot> (<annot
      type="abbreviation">GIFT</annot>) is a method in which oocytes and sperm are
      transferred to one or both fallopian tubes, usually by means of laparoscopically
      directed tubal cannulation. Thus, fertilization occurs in vivo.</feat>
    <feat type="source">BOUROCHE - LACOMBE, A. Les biotechnologies de la reproduction
    chez les mammifères et l'homme. (2001) Vocabulaire français-anglais, INRA Editions,
    Paris, 118 p., ISBN 2-7380-0935-2 </feat>
  </brack>
</struct>
<struct type="termSection" xml:id="BV.203402.TS.6">
  <feat type="term">Gamete intrafallopian transfer</feat>
  <brack>
    <feat type="originatingInstitution">NLM</feat>
    <feat type="originatingDatabaseName">MESH</feat>
  </brack>
  <brack>
    <feat type="originatingInstitution">INIST</feat>
    <feat type="originatingDatabaseName">Vocabulaire multidisciplinaire PASCAL</feat>
  </brack>
  <brack>
    <feat type="originatingInstitution">INRA</feat>
    <feat type="originatingDatabaseName">Biotechnologie de la reproduction</feat>
  </brack>
</struct>
...
</struct>
```

Figure 5: Gamete Intrafallopian Transfer in GMT

## 5. DISCUSSION

The TermSciences initiative deals with the construction of a multi-purpose and multi-lingual terminological database from various source vocabularies produced by major French research institutions. The first requirement of this work was the use of a model that allows for good modeling of data present in these source vocabularies. This was achieved using a data model based on the ISO 16642 standard which was found to be very suitable for modeling term-centred terminological resources into a concept-oriented system. Transformation of terms into concepts was accompanied by transformation of term relationships into concept relationships, i.e. hierarchical and associative relationships are no more at the term level but at the concept one. Adaptations of the traditional terminology principles (wüsterian) are necessary when dealing with specific terminological resources such as thesauri and indexing vocabularies. Thus, the representation of preferred and non-preferred concepts referring to the same descriptor was achieved by introducing a relation at the level of terms. Non-preferred concepts are introduced in the terminological database as separate records but are linked to the preferred concept by a relation occurring at the level of terms. This relation links a term which corresponds to a synonymous concept in a given thesaurus to the term corresponding to the preferred concept which is labeled as being the descriptor. The organization of concepts relevant for a particular domain varies from one source vocabulary to another depending on the degree of precision needed by each application (Rassinoux et al, 1998). Thus, the hierarchy in the MeSH thesaurus may be simple or multiple presenting a given descriptor in different positions in the hierarchy. Furthermore, hierarchies from different source vocabularies may not map correctly, resulting in conflicting positioning of some concepts in the semantic network. Dealing with this topic can be achieved by a) finding a consensual typology of concepts which is not impossible if the level of detail of the typology is not high or b) by representing multiple typologies, i.e. the hierarchies present in the different source vocabularies and additional typologies further introduced.

### 5.1. Reusing of the terminological database

TermSciences is already available on-line and can be used for querying a bibliographical database or helping translator or linguist in a specific subject field. We are planning to add other free bibliographical databases such as PubMed and others. Using the French and English terms contained in a terminological entry, the query is automatically composed and launched on the specified repository. In addition to the cross-language retrieval of relevant documents and citations, another great advantage of this system is the possibility to search bibliographical databases with terms from alternative thesauri and vocabularies. Indexing and cataloging activities being upstream from information retrieval, the terminological database is intended to be connected to bibliographical databases production systems. These systems are those of the TermSciences partners whose needs are about the improvement of their controlled vocabularies management processes and tools, and of the optimization of the indexing process, especially machine-aided indexing programs which performance relies on the quality of the terminological content. The multiple representations (terms) of a given concept which are documented in the terminological database and the variant forms that can be obtained using natural language processing techniques (see bellow) are expected to enhance precision of the machine-aided indexing procedures through consistent interpreting of texts and suggestion of appropriate indexing terms. Another important application is the HAL (Hyper Article en Ligne) institutional open archive of the French researchers which provides authors with an interface enabling them to deposit and index their scientific articles in this repository which is managed by the Center for Direct Scientific Communication, a service unit of the CNRS. At least, this resource will be freely available.

## 5.2. Adding linguistic resources

Additional resources are crucial for a) harmonising the quality and the granularity of the various linguistic descriptions of terms, and b) for purposes such as semi-automatic indexing, information retrieval, translation, etc (Cabre and al., 2005). Natural language processing using on the available lexical features of terms is needed to enhance the recognition rate and quality.

The adding of lexical features in the TermSciences terminological database is being examined from two points of view: tagging of terminological database terms or capturing of lexical features from existing lexical resources such as Morphalou (ATILF) for French terms.

Adding of lexical information is intended to meet another requirement, i.e. to increase the consistency of the set of synonym terms present in a terminological entry. That is, in controlled vocabularies such as those used to build the TermSciences terminological database, morphological variants of the same term are often present and are considered as being synonymous of the preferred term (Zweigenbaum et al. 2003). This results in an artificial inflation of permuted or inflected expressions in some terminological entries. For instance, the MeSH thesaurus presents permuted forms in records such as 'Primary Parkinsonism' and 'Parkinsonism, Primary'. Term tagging or coupling with lexical resources will result in a deflation of the set of terms by discarding the terms which correspond to lexical variants differing from each other only by spelling, word order, number, etc.

In the biomedical field, a salient project, i.e. the Unified Medical Language System (UMLS; McCray et al. 1993) deals with this topic. In this project, lexical knowledge is provided as a distinct source, the SPECIALIST lexicon (McCray, 1998). Coverage of this knowledge source includes both commonly occurring English words and biomedical vocabulary. As English language part in UMLS knowledge sources is greater than that of other languages including French, two projects, i.e. the Unified Medical Lexicon for French (UMLF) which aims at providing a French equivalent for the SPECIALIST lexicon (Zweigenbaum et al. 2003, Zweigenbaum and Grabar, 2003), and the VUMeF project (French Unified Medical Vocabulary) which aims at extending the French part of the UMLS metathesaurus (Darmoni et al.2003).

## 5.3. Corpora

The use of selected corpora represents another important topic for the capture of additional elements in the terminological database such as contexts of use and for terminological extraction. For instance, contexts of use are very useful to translators since they reflect the actual use (or misuse) of a term. The automatic capture of contexts from bibliographical database abstracts or full-text records produced by TermSciences partners is explored as a first step toward context assignment to each term in the terminological database. As human indexers handle the terminological material during rule editing for machine-aided indexing, automatically-captured candidate contexts for terms will be suggested and then verified by human indexers for final selection before addition to the terminological database.

Concerning term extraction, corpora stored in bibliographical databases or incoming bibliographical records subjected to machine-aided indexing routines will be used to suggest candidate terms and candidate semantic relationships between terms (Jacquemin, 1997). The expression of term relationships in texts being revealed by connective words such as 'is called' 'is a', etc (Jacquemin and Bourigault, 2003), cue words and rules for different knowledge domains must be defined through linguistic studies of text samples and then used by computer prohrams to explore these texts and find semantically related terms. Other methods do not require patterns or rules and may use collocation, i.e. cohesive lexical clusters, retrieving (Smadja, 1993)

## 6. Conclusion

The TermSciences initiative aims at building a terminological database by integrating various vocabularies mainly used for indexing purposes. As a first step toward integration, standardization of the source vocabularies was obtained through deployment of the ISO 16642 also called TMF. Although, this standard turned on be suitable for modeling and sharing of the source vocabularies, adaptations were necessary for modeling specific relations which occur frequently in indexing controlled vocabularies, i.e. relations linking non-preferred terms (non-descriptors) to the preferred term (descriptor). Further work is also needed to improve the content of the terminological database and to introduce additional data such as linguistic features, contexts of use, etc.